\documentclass[letterpaper]{article} 
\usepackage{aaai2026}
\usepackage{times}  
\usepackage{helvet}  
\usepackage{courier}  
\usepackage[hyphens]{url}  
\usepackage{graphicx} 
\usepackage{natbib}  
\usepackage{caption} 
\usepackage{algorithm}
\usepackage{algorithmic}
\usepackage{booktabs}
\usepackage{multirow}
\usepackage{amssymb}
\usepackage{hyperref}
\usepackage{newfloat}
\usepackage{listings}
\DeclareCaptionStyle{ruled}{labelfont=normalfont,labelsep=colon,strut=off} 
\floatstyle{ruled}
\newfloat{listing}{tb}{lst}{}
\floatname{listing}{Listing}
\title{SkySplat: Generalizable 3D Gaussian Splatting from Multi-Temporal Sparse Satellite Images}
\author {
    Xuejun Huang\textsuperscript{\rm 1},
    Xinyi Liu\textsuperscript{\rm 1,2 \corrauthor},
    Yi Wan\textsuperscript{\rm 1,2},
    Zhi Zheng\textsuperscript{\rm 3},
    Bin Zhang\textsuperscript{\rm 4},\\
    Mingtao Xiong\textsuperscript{\rm 1},
    Yingying Pei\textsuperscript{\rm 1},
    Yongjun Zhang\textsuperscript{\rm 1,2 \corrauthor}
}
\affiliations {
\textsuperscript{\rm 1}School of Remote Sensing and Information Engineering, Wuhan University, Wuhan, China \\
\textsuperscript{\rm 2}Technology Innovation Center for Collaborative Applications of Natural Resources Data in GBA, Ministry of Natural Resources, Guangzhou, China \\
\textsuperscript{\rm 3}Department of Geography and Resource Management, The Chinese University of Hong Kong, Hong Kong, China \\
\textsuperscript{\rm 4}China Railway Siyuan Survey and Design Group Co., LTD, Wuhan, China \\
    {\small\texttt{liuxy0319@whu.edu.cn, zhangyj@whu.edu.cn}}
    }

\usepackage{bibentry}
\usepackage{amsmath}  
\usepackage{placeins} 

\begin{document}
\nocopyright \captionsetup[table]{name=Tab.}

\maketitle
\begin{figure*}[t]
    \centering
    \includegraphics[width=0.98\textwidth]{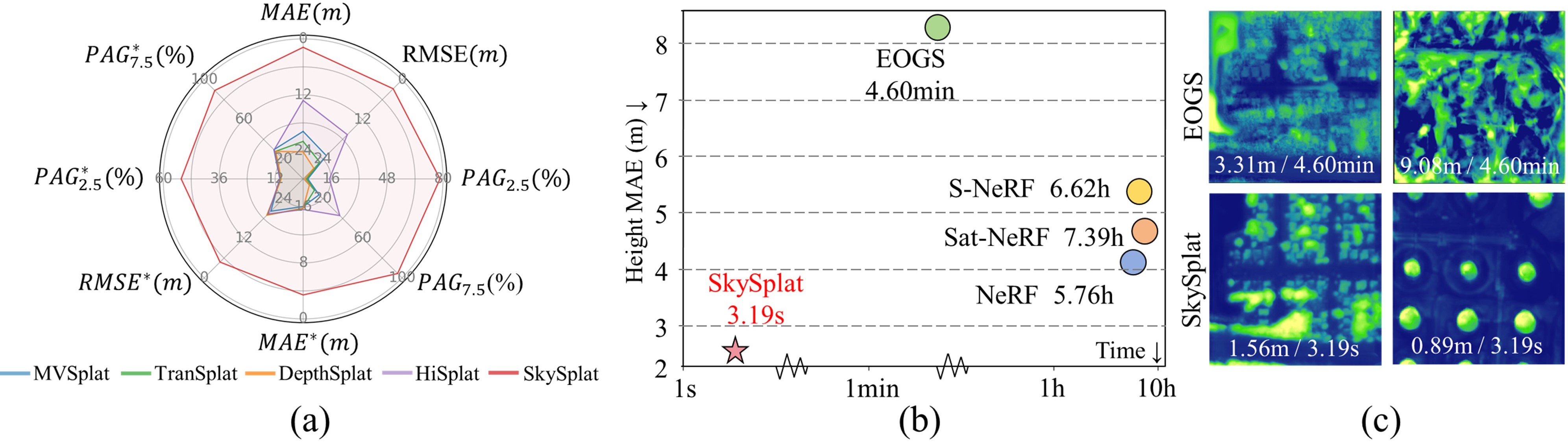}
    \caption{
        \textbf{Comparison with existing methods}. (a) Generalization results on the DFC19 \cite{bosch2019semantic,le20192019} and MVS3D \cite{bosch2016multiple} datasets. Results from the MVS3D dataset are marked with *. SkySplat achieves the best performance among all competitors. (b) Per-scene optimization results. SkySplat reaches optimal performance in just 3.19 seconds. (c) DSM reconstruction results on two areas of interest (AOIs). We report both MAE and reconstruction time.
    }
    \label{fig:teaser}
\end{figure*}

\begin{abstract}
Three-dimensional scene reconstruction from sparse-view satellite images is a long-standing and challenging task. While 3D Gaussian Splatting (3DGS) and its variants have recently attracted attention for its high efficiency, existing methods remain unsuitable for satellite images due to incompatibility with rational polynomial coefficient (RPC) models and limited generalization capability. Recent advances in generalizable 3DGS approaches show potential, but they perform poorly on multi-temporal sparse satellite images due to limited geometric constraints, transient objects, and radiometric inconsistencies. To address these limitations, we propose SkySplat, a novel self-supervised framework that integrates the RPC model into the generalizable 3DGS pipeline, enabling more effective use of sparse geometric cues for improved reconstruction. SkySplat relies only on RGB images and radiometric-robust relative height supervision, thereby eliminating the need for ground-truth height maps. Key components include a Cross-Self Consistency Module (CSCM), which mitigates transient object interference via consistency-based masking, and a multi-view consistency aggregation strategy that refines reconstruction results. Compared to per-scene optimization methods, SkySplat achieves an 86 times speedup over EOGS with higher accuracy. It also outperforms generalizable 3DGS baselines, reducing MAE from 13.18 m to 1.80 m on the DFC19 dataset significantly, and demonstrates strong cross-dataset generalization on the MVS3D benchmark. The code is publicly available at \href{https://github.com/NanCheng2001/SkySplat-main}{https://github.com/NanCheng2001/SkySplat-main}
\end{abstract}


\section{Introduction}
Three-dimensional scene reconstruction from sparse-view satellite images remains a fundamental challenge in photogrammetry and computer vision. This technique has a wide range of applications such as digital twins and urban planning \cite{zheng2024digital}.

Recent advances in multi-view stereo (MVS) methods have demonstrated the potential of deep neural networks for accurate 3D reconstruction from satellite images \cite{gao2021rational,gao2023general}. However, these methods typically rely on fully supervised learning and often degrade on multi-temporal satellite images due to radiometric and seasonal variations. In contrast, NeRF-based methods \cite{mildenhall2021nerf,barron2021mip,barron2022mip} have gained attention for 3D reconstruction without ground-truth height supervision. Variants like S-NeRF \cite{derksen2021shadow} and Sat-NeRF \cite{mari2022sat} have been proposed to address the challenges posed by multi-temporal satellite images. However, these methods are computationally expensive and require long training times \cite{charatan2024pixelsplat}. 

To improve efficiency, the original 3D Gaussian Splatting (3DGS) \cite{kerbl20233d} has emerged as a novel 3D representation, enabling fast rendering. Several extensions \cite{aira2025gaussian,bai2025satgs} for satellite images have been proposed to enhance reconstruction quality and training efficiency. However, they still require several minutes to reconstruct a 256 × 256 m² scene through per-scene optimization and rely on a large number of input views.

More recently, data-driven generalizable 3DGS methods have been developed to improve transferability and reconstruction speed. These approaches use neural networks to directly infer per-pixel Gaussian splatting parameters for unseen scenes \cite{xu2025depthsplat}. Representative examples include MVSplat \cite{chen2024mvsplat}, TranSplat \cite{zhang2025transplat}, and HiSplat \cite{tang2024hisplat}, which achieve efficient 3D reconstruction from sparse-view images in a single feed-forward pass. However, these methods are not directly applicable to multi-temporal satellite images due to several challenges: (1) the unique pushbroom imaging mode of satellites, which violates the standard pinhole camera assumption; (2) interference from transient objects, such as moving vehicles and vegetation changes; and (3) radiometric inconsistencies across multi-temporal images caused by variations in illumination and atmospheric conditions.

To address these challenges, we propose SkySplat, a novel self-supervised framework for generalizable 3D reconstruction from multi-temporal sparse satellite images. SkySplat is the first to explicitly incorporate the satellite-specific rational polynomial coefficient (RPC) model into the generalizable 3DGS pipeline, enabling accurate scene reconstruction without pinhole approximations. Extensive experiments show that SkySplat achieves significantly higher accuracy than previous generalizable methods while being 86× faster than SOTA per-scene optimization approaches (see Figure \ref{fig:teaser}). Our main contributions can be summarized as follows:

\begin{itemize}
\item We propose SkySplat, the first generalizable 3DGS framework that incorporates the RPC model without approximation, achieving accurate scene reconstruction and an 86× speedup over the SOTA per-scene optimization method EOGS.
\item We design a Cross-Self Consistency Module (CSCM) to minimize the impact of transient objects during training, and incorporate monocular relative heights as additional supervision to address radiometric inconsistencies in multi-temporal satellite images.
\item We introduce a multi-view consistency aggregation strategy to further enhance reconstruction accuracy.
\end{itemize}

\section{Related Work}

\subsection{NeRF and 3DGS for Satellite Images}
In recent years, NeRF \cite{mildenhall2021nerf} and 3DGS \cite{kerbl20233d} have made significant progress in 3D scene reconstruction. However, the unique characteristics of satellite images—such as the RPC model and radiometric inconsistencies—pose substantial challenges for reconstruction. To address these issues, several methods \cite{derksen2021shadow,mari2022sat,mari2023multi} have extended NeRF to satellite domains by improving light transport models, ray sampling strategies based on the RPC model, and incorporating shadow modeling techniques. Subsequent efforts further enhanced NeRF-based scene reconstruction by introducing geometric constraints or priors \cite{behari2024sundial,liu2025sat}. In contrast, more recent approaches like EOGS \cite{aira2025gaussian} and SatGS \cite{bai2025satgs} leverage the real-time and efficient rendering capabilities of 3DGS, adapting it to multi-temporal satellite images for more efficient photogrammetry. Despite these advances, current methods still struggle with limited generalization and often require more than ten input views to achieve satisfactory performance.

\subsection{Sparse-View Scene Reconstruction}
The insufficient geometric constraints between sparse-view images poses significant challenges for scene reconstruction \cite{shi2024zerorf}. In computer vision, existing sparse-view methods can be categorized into two groups: per-scene reconstruction methods and generalizable reconstruction methods \cite{zhang2025transplat}. The former typically leverage multi-view geometric constraints \cite{truong2023sparf,deng2022depth}, or incorporate stronger supervision from pre-trained models to improve reconstruction quality \cite{zhu2024fsgs,yu2022monosdf}. However, these methods are often hindered by time-consuming optimization processes, leading to low efficiency. In contrast, generalizable methods perform scene reconstruction in a single feed-forward pass, demonstrating strong generalization by learning powerful priors from large-scale datasets \cite{yu2021pixelnerf,yang2023freenerf,liu2024mvsgaussian}. However, in the context of sparse-view satellite images, existing reconstruction methods either rely on ground truth height supervision \cite{gao2021rational} or require per-scene optimization, which is computationally expensive \cite{zhang2024fvmd}.

\subsection{Generalizable 3DGS}
Generalizable 3DGS models have recently gained significant attention due to their high efficiency and strong generalization capabilities. PixelSplat \cite{charatan2024pixelsplat} pioneers this approach by leveraging transformer-encoded features to predict Gaussian parameters directly. Subsequent works \cite{chen2024mvsplat,zhang2025transplat,tang2024hisplat} typically employ depth prediction networks to regress 3D Gaussians, further enhancing generalization. However, due to transient objects, radiometric inconsistencies, and the unique RPC model, directly applying them to multi-temporal satellite images results in suboptimal performance. To address this, we explicitly integrate the RPC model into the generalizable 3DGS pipeline. Our method enables accurate scene reconstruction in a self-supervised manner by filtering transient objects and incorporating relative height supervision.

\begin{figure*}[h!]
    \centering
    \includegraphics[width=0.98\linewidth]{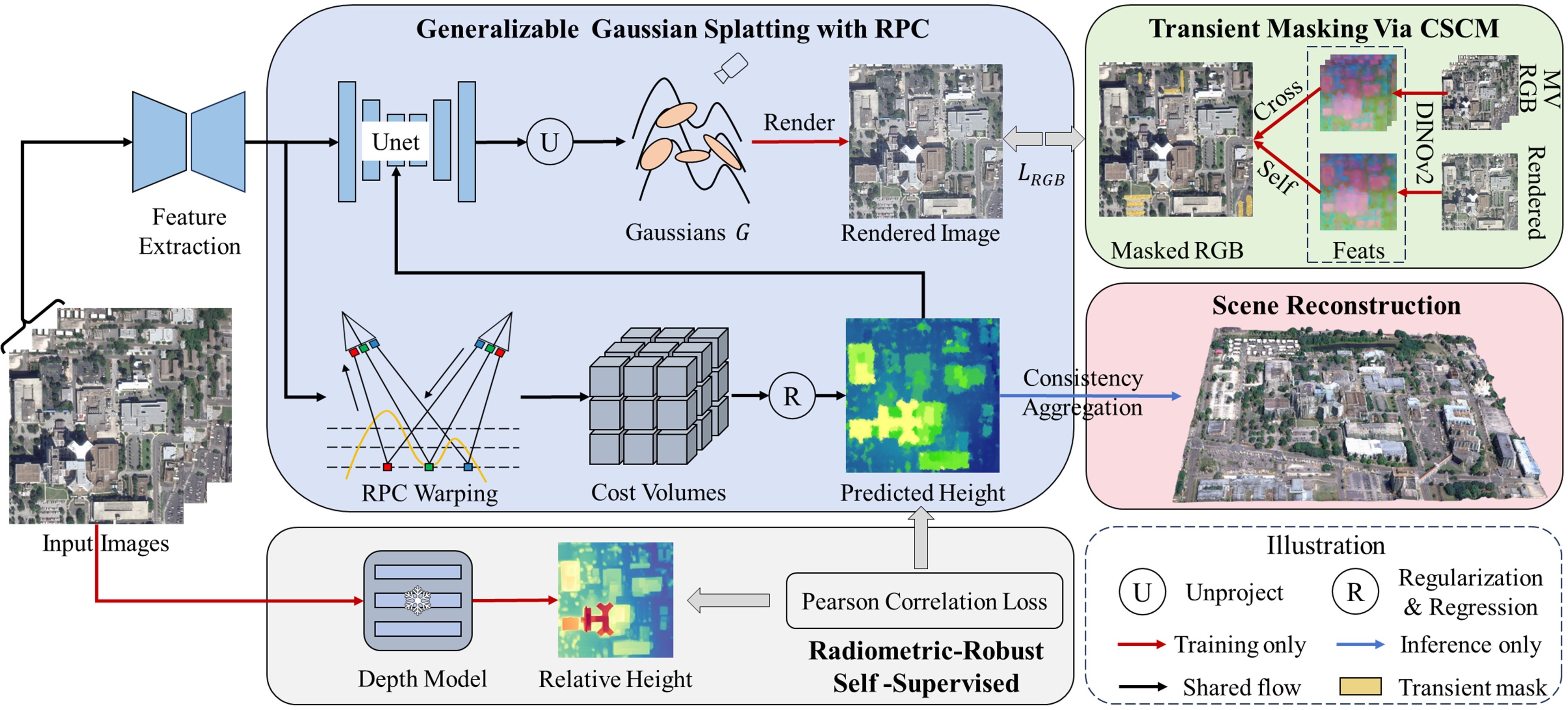}
    \caption{\textbf{Overview of the proposed SkySplat.} The depth model is from \cite{yang2024depth}, Depth Anything V2 (DAMV2).
    }
    \label{fig:framework}
\end{figure*}

\section{Method}
Given a set of \(N\) sparse-view satellite images \(\left\{I_{i}\right\}_{i=1}^{N}\) and their corresponding RPCs, we aim to learn a generalizable model that can accurately reconstruct 3D scenes without requiring ground-truth height supervision. To achieve this goal, we propose SkySplat, a feed-forward framework that avoids per-scene optimization. It first matches and fuses geometric cues from the images with the RPC model to predict height and Gaussian parameters. Then, transient masks are generated by the CSCM to minimize the impact of dynamic objects during training. Next, monocular relative heights from a pretrained depth model \cite{yang2024depth} are incorporated as additional supervision to handle radiometric inconsistencies in multi-temporal satellite images. Finally, a consistency aggregation strategy is applied to refine the scene reconstruction results. An overview of the proposed SkySplat is shown in  Figure \ref{fig:framework}.

\subsection{Generalizable Gaussian Splatting with RPC}
The generalizable 3DGS with RPC involves four main steps: multi-view feature extraction, RPC-guided cost volume construction, height estimation and Gaussian parameter prediction.
\subsubsection{Multi-View Feature Extraction.} To ensure efficiency, we follow the feature extraction strategy used in MVSPlat \cite{chen2024mvsplat}. Specifically, we avoid any 3D convolutions and use a multi-view Transformer to aggregate features across different views, which produces features \(\{F_i\}_{i=1}^{N}\).
\subsubsection{RPC-Guided Cost Volume Construction.} Next, we compute feature correlations across views using the differentiable rpc warping to construct cost volumes  \(\{C_i\}_{i=1}^{N}\) \cite{gao2021rational}. For each reference feature \(F_i\)
, we sample \(M\) height candidates \(\{h_m\}_{m=1}^{M}\) within a predefined elevation range. Using inverse RPC projection, each pixel coordinates \((u_i, v_i)\) of \(F_i\) is back-projected with height \(h_m\) to obtain corresponding 3D coordinates:
\begin{equation}
(\text{Lat}_{i}^{m}, \text{Lon}_{i}^{m}, \text{Hei}_{i}^{m}) = \text{RPC}_{\text{ref}}^{-1}(u_i, v_i, h_m)  \tag{1}
\end{equation}
These 3D points are then projected into each source view. We sample source features via interpolation at the projected positions and warp them to the reference view:
\begin{equation}
F_{j \rightarrow i}^{h_m} = \text{Inter}(F_j, \text{RPC}_{\text{src}}(\text{Lat}_{i}^{m}, \text{Lon}_{i}^{m}, \text{Hei}_{i}^{m})) \tag{2}
\end{equation}
where \(Inter\) refers to the interpolation operation. A variance-based operation \cite{gao2021rational} is then applied across the warped features to compute the final cost volumes \(\{C_i\}_{i=1}^{N}\).
\subsubsection{Height Estimation.} Each regularized cost volume is then used to estimate height. We apply a \(soft\) \(argmin\) operation along the height dimension to produce a per-view height map, denoted as \(\{\hat{h}_i\}_{i=1}^N
\).
\subsubsection{Gaussian Parameter Prediction.} With the estimated heights, we compute the 3D center positions \(\mu_{3D}\) of the Gaussians by inverse RPC projection followed by coordinate correction. Then, we adopt a 2D U-Net, following MVSPlat \cite{chen2024mvsplat}, to predict the remaining Gaussian parameters. These include the scaling factor \(S\), rotation quaternion \(R\), spherical harmonic coefficients \(C\), and opacity \(\alpha\):
\begin{equation}
G = \{\mu_{3D}, S, R, C, \alpha\}
\tag{3}
\end{equation}
Note that the rotation quaternion \(R\) is derived by approximating the RPC model with the pinhole camera model during training \cite{zhang2019leveraging}. Since training is conducted on 256×256 patches, the introduced error is negligible. During inference, we use the exact \(\mu_{3D}\) obtained from the RPC projection, avoiding any approximation error.

\begin{figure}[!t]
\centering
\includegraphics[width=\linewidth]{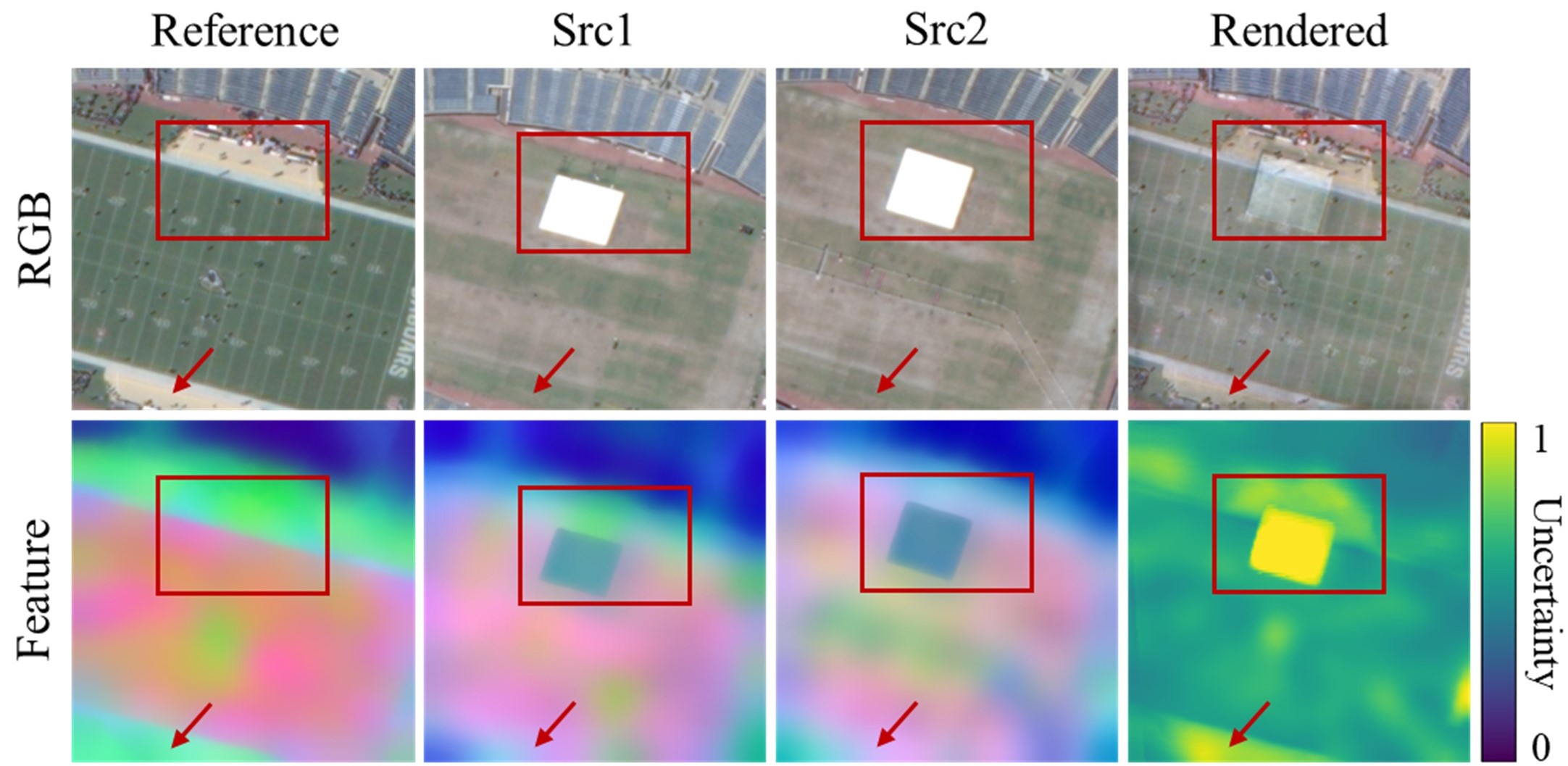}
\caption{\textbf{Visualization of uncertainty regions generated by CSCM.} Top: Three-views input images and the rendered reference image. Bottom: DINOv2 feature visualizations (first three columns) via PCA \cite{abdi2010principal}, and uncertainty map (last column) of the reference view. Red boxes and arrows highlight transient objects,  where gradient propagation is halted.
}
\label{fig:trans}
\end{figure}

\subsection{Transient Masking via Cross-Self Consistency}
Current methods for handling transient objects in multi-temporal satellite images often rely on per-scene optimization, limiting their generalization \cite{bao2024distractor}. To address this issue, we propose the Cross-Self Consistency Module (CSCM), a transient masking method based on both cross-view and self-view feature similarity. As shown in Figure \ref{fig:trans}, CSCM automatically identifies uncertain regions during training, where the RGB image supervision loss is halted.

Considering the photometric stability of the features \cite{liu2025sem}, we extract feature maps \(\{feat_i\}_{i=1}^{N}\) using DINOv2 with FeatUp \cite{oquab2023dinov2,fu2024featup}. Following Equations (1) and (2), we build feature correspondences across views to obtain projected features \(\left\{ \text{feat}_{j \rightarrow i} \right\}_{i=1}^{N}
\).
At pixels selected by multi-view geometric consistency filters (see \textit{Scene Reconstruction via Consistency Aggregation}), we compute the cosine similarity between \(feat_i\) and \({feat}_{j\rightarrow i}\), then convert it into a confidence map in [0,1] \cite{kulhanek2024wildgaussians}:
\begin{equation}
Q_{cv} = \max\left(2 \cdot \cos(\text{feat}_i, \text{feat}_{j \rightarrow i}) - 1,\, 0\right)
\tag{4}
\end{equation}
where \(Q_{cv}\) measures the features consistency across views. Similarities below 0.5 yield zero confidence. To handle invalid regions caused by geometric filtering, we introduce a self-view confidence map \(Q_{sv}\), computed based on the similarity between the reference features \(feat_i\) and the rendered features \(feat_i^\prime\) \cite{fu2025robustsplat}:
\begin{equation}
Q_{sv} = \max\left(2 \cdot \cos(\text{feat}_i, \text{feat}_i^\prime) - 1,\, 0\right)
\tag{5}
\end{equation}
As \(Q_{sv}\) is usually lower than \(Q_{cv}\), we calibrate its scale using the mean ratio over valid regions of \(Q_{cv}\). The final confidence map \(Q\) is constructed by replacing the invalid regions in \(Q_{cv}\) with the calibrated \(Q_{sv}\). Finally, \(Q\) is thresholded at \(\tau=0.2\) to produce a binary mask \(M\), which suppresses misleading supervision from RGB images in regions affected by transient objects. The module is applied after 35k iterations, as height estimates become more reliable. See supplementary for hyperparameter analysis.

\subsection{Radiometric-Robust Self-Supervised Learning}
To mitigate suboptimal local minima caused by varying imaging conditions in images supervision \cite{zhang2024satensorf}, we introduce auxiliary supervision based on relative height, which is more robust to illumination changes \cite{liu2025sat}.

We first obtain relative height maps \(\{H_i\}_{i=1}^{N}\) from Depth Anything V2 (DAMV2) \cite{yang2024depth} for each view. To address the scale ambiguity issue, we follow previous approach \cite{zhu2024fsgs} and supervise the predicted absolute height maps \(\{\hat{h}_i\}_{i=1}^N\) using the Pearson correlation loss:
\begin{equation}
\mathcal{L}_{\text{hei}} = \frac{\text{Cov}(H, \hat{h})}{\sqrt{\text{Var}(H) \cdot \text{Var}(\hat{h})}} \tag{6}
\end{equation}
where $\text{Cov}(\cdot, \cdot)$ and $\text{Var}(\cdot)$ denote covariance and variance, respectively. Unlike previous work \cite{liu2025sat}, SkySplat does not require explicit scale alignment, as it directly captures similarity between height distributions.

In addition, we apply both LPIPS and MSE losses between rendered and ground-truth images, guided by the mask \(M\), where 1 indicates stable (non-transient) regions:
\begin{equation}
\mathcal{L}_{\text{rgb}} = M \odot \text{LPIPS}(I_{\text{render}}, I_{\text{gt}}) + M \odot (I_{\text{render}} - I_{\text{gt}})^2 \tag{7}
\end{equation}
where \(\odot\) denotes element-wise multiplication. The rendered image \(I_{\text{render}}\) is generated by native 3DGS rendering, where the RPC model is approximated by the pinhole camera model \cite{zhang2019leveraging}:
\begin{equation}
I_{\text{render}} = \sum_i c_i \cdot \alpha_i \prod_{j=1}^{i-1} (1 - \alpha_j) \tag{8}
\end{equation}
As mentioned earlier, the approximation error is negligible due to small training patches, and no rendering is used during inference. The final self-supervised loss is defined as: $\mathcal{L} = \mathcal{L}_{\text{rgb}} + \mathcal{L}_{\text{hei}}$.

\subsection{Scene Reconstruction via Consistency Aggregation}
To enhance reconstruction accuracy, we adopt a Multi-view Consistency Aggregation strategy inspired by \cite{liu2024mvsgaussian,gao2021rational}, which filters out noisy Gaussian points with high reprojection errors across views. 

Specifically, given a predicted height map \(\hat{h}_i
\) for the reference view and \(\hat{h}_j
\) for the source view, we first project a point \(p\) from the reference view to the source view using Equations (1) and (2), obtaining the projection point \(q\). The source-view height \(\hat{h}_j(q)
\) is then sampled at this location. Subsequently, the point \(q\) is  reprojected back to the reference view to obtain \(p^{\prime}\), using: 
\begin{equation}
p^{\prime} = \text{RPC}_{\text{ref}} \left( \text{RPC}_{\text{src}}^{-1}(u_q, v_q, \hat{h}_j(q)) \right) \tag{9}
\end{equation}
where \((u_q,v_q)\) denote the pixel coordinates of \(q\). Then, we get  the height \(\hat{h}_i\left(p^{\prime}\right)
\), and compute the geometric and height reprojection errors as:
\begin{equation}
\delta_p = \| p - p^{\prime} \|_2 \tag{10}
\end{equation}
\begin{equation}
\delta_h = \frac{|\hat{h}_i(p) - \hat{h}_i(p^{\prime})|}{|\hat{h}_i(p)|} \tag{11}
\end{equation}
Only points with $\delta_p < 3$ and $\delta_h < 0.2$ are retained as reliable 3D points. Finally, these filtered points are orthogonally projected onto a regular 2D grid. For each grid cell, we keep the maximum height among all assigned points to generate the final DSM.

\section{EXPERIMENTS}

\begin{table*}[!t]
\centering
\setlength{\tabcolsep}{1.3mm} 
\begin{tabular}{l|cccc|cccc}
\toprule
\multirow{2}{*}{\textbf{Method}} & \multicolumn{4}{c|}{DFC19 Dataset} & \multicolumn{4}{c}{MVS3D Dataset} \\
 & MAE(m)$\downarrow$ & RMSE(m)$\downarrow$ & $\text{PAG}_{2.5}$(\%)$\uparrow$ & $\text{PAG}_{7.5}$(\%)$\uparrow$ & MAE(m)$\downarrow$ & RMSE(m)$\downarrow$ & $\text{PAG}_{2.5}$(\%)$\uparrow$ & $\text{PAG}_{7.5}$(\%)$\uparrow$ \\
\midrule
pixelSplat         & 176.03 & 189.26 & 0.02  & 0.07  & 29.15 & 38.41 & 6.18  & 18.36 \\
MVSplat            & 19.82  & 22.96  & 2.71  & 16.11 & 16.05 & 20.20 & 9.99  & 29.54 \\
TranSplat          & 21.96  & 24.94  & 1.86  & 11.40 & 15.81 & 19.01 & 9.44  & 27.85 \\
DepthSplat         & 24.21  & 26.76  & 0.89  & 6.33  & 15.81 & 19.01 & 9.44  & 27.85 \\
HiSplat            & 13.18  & 16.59  & 15.06 & 37.08 & 15.63 & 19.37 & 9.96  & 29.34 \\
\midrule
SkySplat w/o C.A.      & \underline{2.07}  & \underline{3.07}  & \underline{75.08} & \underline{94.91} & \underline{3.48} & \underline{4.80} & \underline{50.76} & \underline{89.16} \\
SkySplat          & \textbf{1.80} & \textbf{2.68} & \textbf{78.27} & \textbf{95.57} & \textbf{3.42} & \textbf{4.79} & \textbf{52.32} & \textbf{89.35} \\
\bottomrule
\end{tabular}
\caption{\textbf{Quantitative comparison with generalizable methods on both datasets.} 
SkySplat achieves the best overall performance across all metrics on both datasets. The results of “SkySplat w/o C.A.” demonstrate the contribution of our core framework even without the consistency aggregation strategy. (\textbf{Bold} indicates best, \underline{underline} indicates second best.)}
\label{tab:quant_resultsI}
\end{table*}

\begin{table*}[!t]
\centering
\setlength{\tabcolsep}{1.7mm} 
\begin{tabular}{l|ccc|cc|ccc|c}
\toprule
\textbf{Method} & JAX004 & JAX068 & JAX260 & OMA212 & OMA315 & IARPA001 & IARPA002 & IARPA003 & Time \\
\midrule
NeRF & 3.30 & 6.33 & 3.09 & 1.16 & 3.01 & 4.11 & 6.05 & 6.02 & 5.76 h \\
S-NeRF & 3.28 & 7.47 & 4.88 & 3.24 & 2.98 & 4.97 & 9.71 & 6.55 & 6.62 h \\
Sat-NeRF & \underline{3.27} & 6.53 & 5.28 & 3.16 & 2.99 & 4.63 & 6.65 & \underline{4.92} & 7.39 h \\
EOGS & 3.31 & 6.67 & 6.41 & 9.08 & 6.38 & 5.90 & 13.79 & 14.83 & 4.60 min \\
\midrule
SkySplat w/o C.A. & \textbf{1.56} & \underline{4.24} & \underline{2.68} & \underline{0.90} & \underline{1.53} & \underline{3.14} & \underline{3.89} & \textbf{3.41} & \textbf{3.13 s} \\
SkySplat & \textbf{1.56} & \textbf{3.86} & \textbf{2.46} & \textbf{0.89} & \textbf{1.51} & \textbf{3.10} & \textbf{3.75} & \textbf{3.41} & \underline{3.19 s} \\
\bottomrule
\end{tabular}
\caption{\textbf{Quantitative comparison with per-scene optimization methods across three cities.} Reported metrics include MAE on the elevation (meters) and reconstruction time for each method. (\textbf{Bold} indicates best, \underline{underline} indicates second best.)}
\label{tab:all_city_results}
\end{table*}

\subsection{Experimental Setup}
\subsubsection{Datasets.} We train and evaluate our model on the large-scale DFC19 dataset \cite{bosch2019semantic,le20192019}, which contains multi-temporal images from Jacksonville (JAX) and Omaha (OMA) , all with a ground sampling distance (GSD) of 0.3\,m. Following MVSPlat \cite{chen2024mvsplat}, we select three views for each scene and crop them into 256×256 patches, yielding 11,648 training and 1,472 test samples. For cross dataset evaluation, we use three AOIs from the MVS3D dataset \cite{bosch2016multiple}, which share the same GSD, consistent with EOGS \cite{aira2025gaussian}. To compare with per-scene optimization methods, we also evaluate on five AOIs from DFC19 dataset. RPC camera models in both datasets are refined via bundle adjustment.
\subsubsection{Evaluation Metrics.} We evaluate all methods using LiDAR-based DSMs with a GSD of 0.3--0.5\,m. Metrics include mean absolute error (MAE), root mean square error (RMSE), and percentage of accurate grids in total (\(PAG\)) \cite{gao2023general,huang2025mvsr3d}. For example, \(PAG_{2.5}\) represents the ratio of grid cells with an L1 distance error below 2.5\,m, and \(PAG_{7.5}\) represents the ratio for errors below 7.5\,m. Additional metrics and results on novel view synthesis are presented in the Supplements (see Appendix A.2).
\subsubsection{Implementation Details.} We conduct all experiments on a server equipped with eight NVIDIA® GeForce RTX 4090 GPUs (24 GB VRAM each), running on Ubuntu 22.04. All models are trained for 20 epochs using AdamW with a learning rate of $2 \times 10^{-4}$ and a batch size of 3. For each scene, we use a fixed height sampling range \([h_{\min}, h_{\max}]\) with 64 samples, derived from publicly DEMs or LiDAR. In addition, all images of eight AOIs are resized to 768×768 for generalizable 3DGS, and RPCs are adjusted accordingly. No post-processing (e.g., consistency aggregation) is applied during visualization for fair comparison. 

It is worth noting that, to adapt the generalizable 3DGS  compared in Table \ref{tab:quant_resultsI} to satellite images, the RPC model is approximated by the pinhole camera model \cite{zhang2019leveraging}. Due to the significant distance between the satellite camera and the scene, higher numerical precision is employed to prevent numerical instability and overflow issues that may arise from the large depth values.

\begin{figure*}[h!]
    \centering
    \includegraphics[width=0.98\linewidth]{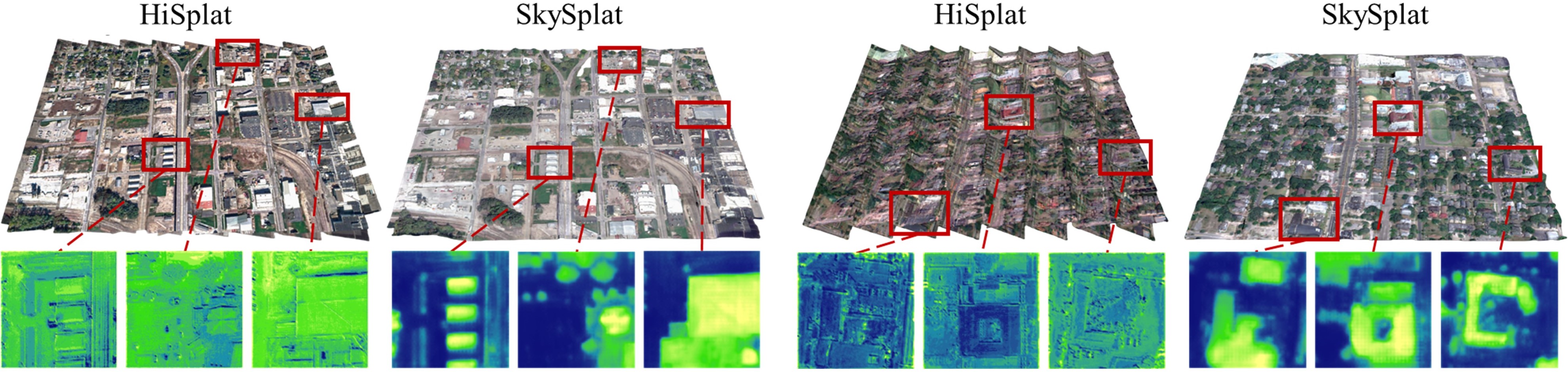}
    \caption{\textbf{Comparisons of 3D Gaussians (top) and height maps (bottom).} SkySplat generates smoother and more accurate results, highlighting its effectiveness.}
    \label{fig:3DGS}
\end{figure*}

\subsection{Generalization Results}
We compare SkySplat with SOTA generalizable methods for sparse-view scene reconstruction. All models are trained on the DFC19 training set, and quantitative results on the DFC19 test set are reported in Table \ref{tab:quant_resultsI}. SkySplat consistently outperforms all baselines across all metrics. Compared to HiSplat, it achieves an 11.38 m lower MAE, a 13.91 m lower RMSE, a 63.21\% higher \(PAG_{2.5}\), and a 58.49\% higher \(PAG_{7.5}\). Notably, even without the consistency aggregation (C.A.) strategy, SkySplat still delivers the best performance, as shown in the second-to-last row of Table \ref{tab:quant_resultsI}.

To further evaluate generalization ability, we directly apply the model trained on the DFC19 to the MVS3D test set. SkySplat again outperforms all baselines, demonstrating strong cross-dataset generalization.

We stitch together 64 non-overlapping outputs, each generated from three 256×256 images, to form a large-scale 3D Gaussian scene. As shown in Figure \ref{fig:3DGS}, SkySplat produces higher-quality 3D Gaussians and more accurate height maps, especially in challenging regions. These results highlight the superiority of our RPC-based framework in achieving accurate 3D reconstructions.

\begin{figure}[!t]
\centering
\includegraphics[width=\linewidth]{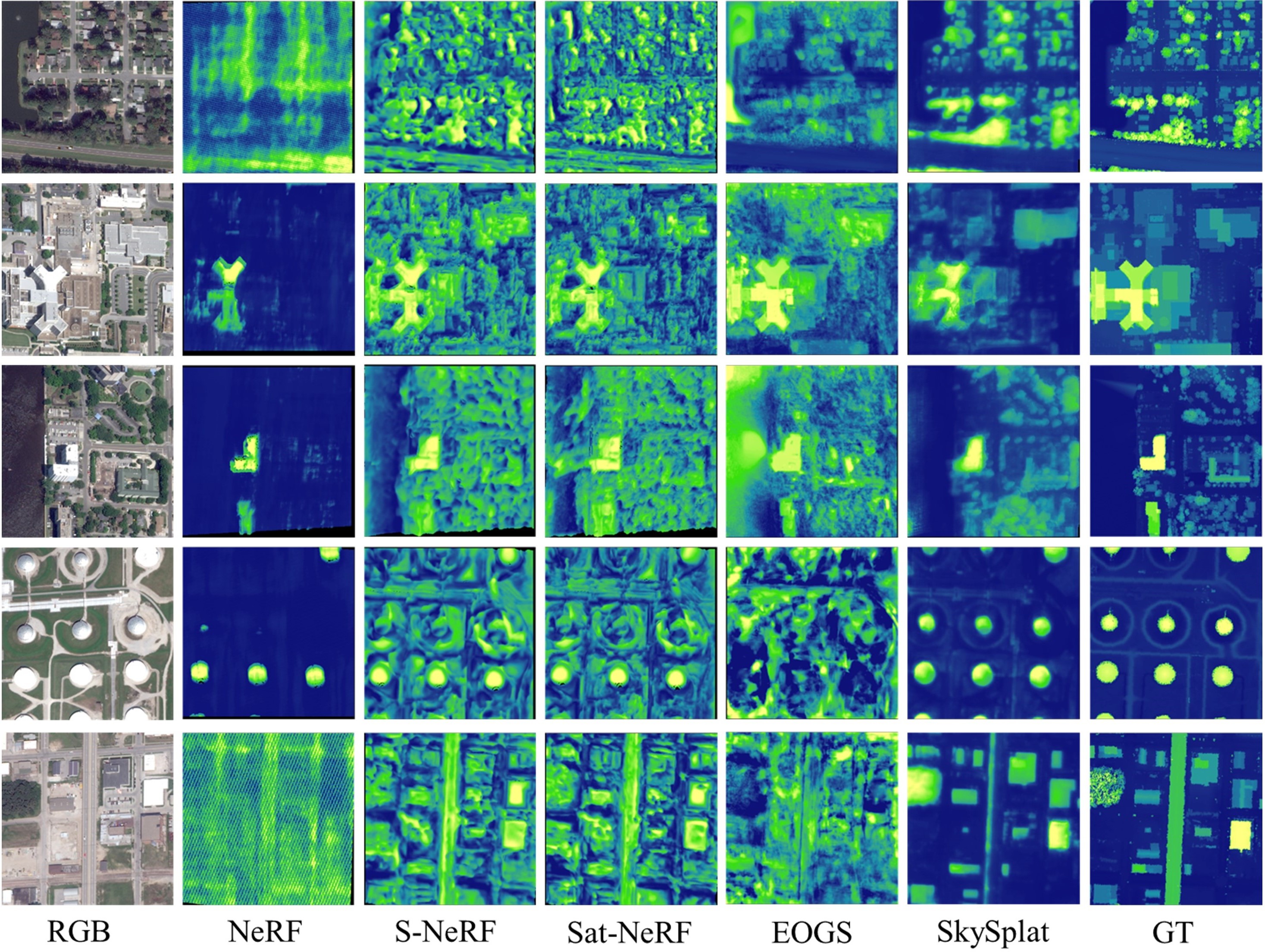}
\caption{\textbf{Predicted DSMs of the DFC19 areas.} From top to bottom: JAX004, JAX068, JAX260, OMA212, OMA315. DSM resolution: 50\,cm/pixel.
}
\label{fig:dsm1}
\end{figure}

\subsection{Per-Scene Optimization Results}
Table \ref{tab:all_city_results} reports quantitative comparisons between SkySplat and SOTA methods that rely on per-scene optimization. Benefiting from strong generalization capability, SkySplat achieves the best performance while drastically reducing reconstruction time. 

Specifically, it reduces the average reconstruction time from several minutes (e.g., 4.60\,min for EOGS) or even hours (e.g., 5–8\,h for NeRF-based methods) to just 3.19 seconds. Despite being nearly 86× faster than EOGS, it still delivers significantly lower MAE across all AOIs. For instance, on IARPA 002 and IARPA 003, SkySplat achieves errors of 3.75\,m and 3.41\,m, compared to 13.79\,m and 14.83\,m by EOGS. Even without the C.A. strategy, our model maintains strong accuracy, outperforming all baselines in both DFC19 and MVS3D datasets.

\begin{figure}[!t]
\centering
\includegraphics[width=\linewidth]{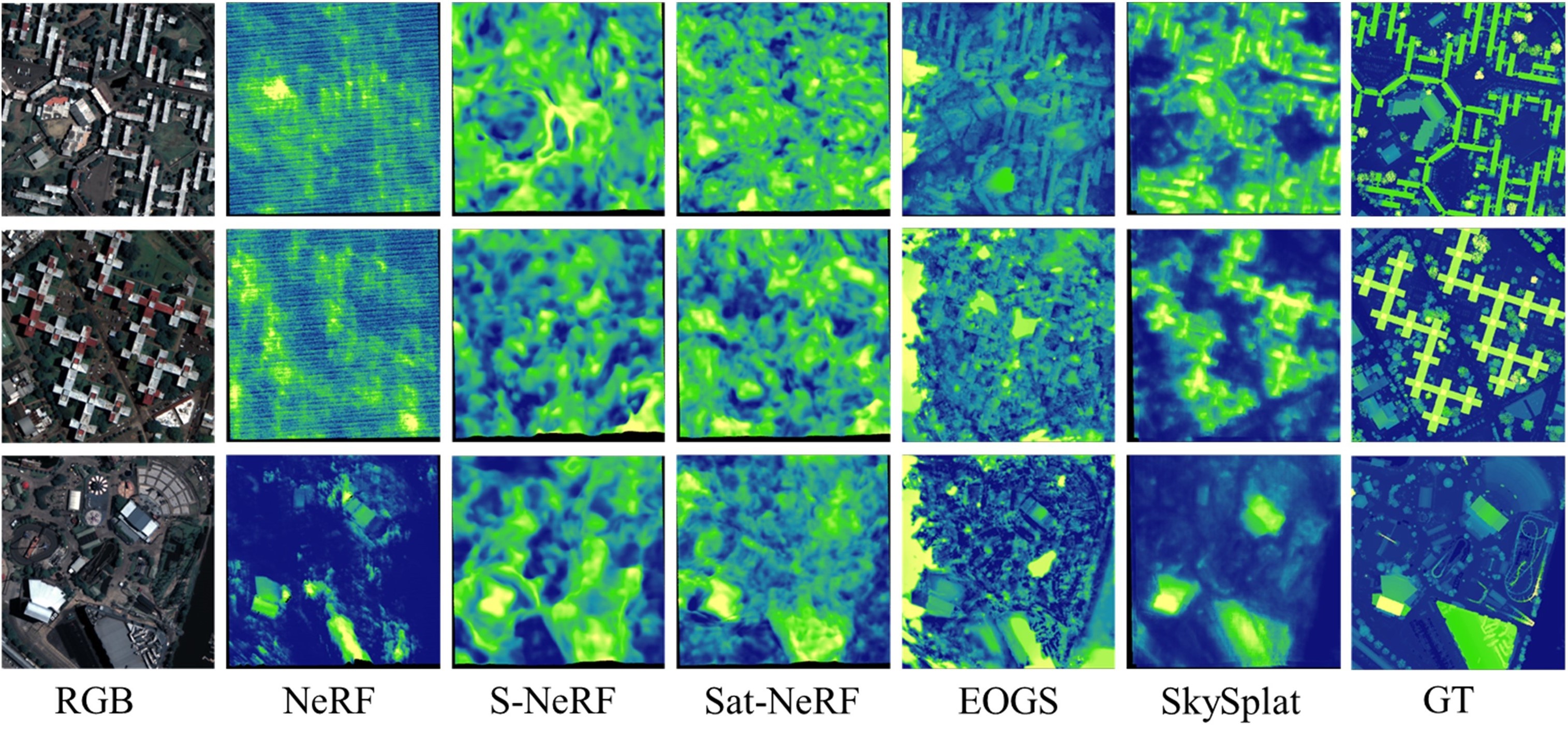}
\caption{\textbf{Predicted DSMs of the MVS3D areas.} From top to bottom: IARPA001, IARPA002, IARPA003. DSM resolution: 30\,cm/pixel.
}
\label{fig:dsm2}
\end{figure}

Visual results in Figures \ref{fig:dsm1} and \ref{fig:dsm2} further highlight the advantages of SkySplat in preserving structural details. Even without fine-tuning, it consistently produces more accurate and coherent 3D reconstructions compared to existing methods. Additionally, the average MAE and reconstruction time of each method are visualized in Figure \ref{fig:teaser} (b).

\begin{table*}[!t]
\centering
\setlength{\tabcolsep}{2.8mm} 
\begin{tabular}{c|c|c|cccc}
\toprule
CSCM & R.H.S. & C.A. & MAE\,(m)$\downarrow$ & RMSE\,(m)$\downarrow$ & $\mathbf{PAG}_{2.5}$\,(\%)$\uparrow$ & $\mathbf{PAG}_{7.5}$\,(\%)$\uparrow$ \\
\midrule
 &  &  & 6.07 & 7.33 & 29.46 & 68.57 \\
\checkmark &  &  & 5.94 & 7.19 & 29.46 & 70.37 \\
 & \checkmark &  & 2.25 & 3.30 & 72.90 & 94.00 \\
\checkmark & \checkmark &  & 2.07 & 3.07 & 75.08 & 94.91 \\
\checkmark & \checkmark & \checkmark & \textbf{1.80} & \textbf{2.68} & \textbf{78.27} & \textbf{95.57} \\
\bottomrule
\end{tabular}
\caption{\textbf{Ablation study on the DFC19 dataset.} We evaluate the contributions of CSCM (Cross-Self Consistency Module), R.H.S. (Relative Height Supervision), and C.A. (Consistency Aggregation). (\textbf{Bold} indicates best.)}
\label{tab:ablation_df19}
\end{table*}

\begin{table*}[!t]
\centering
\setlength{\tabcolsep}{2.8mm} 
\begin{tabular}{c|l|cccc}
\toprule
\textbf{Views} & \textbf{Method} & JAX068 & JAX260 & OMA212 & OMA315 \\
\midrule
\multirow{2}{*}{5} 
& S2P (MGM)$^{\ast}$ & 2.35 & 3.29 & 1.50 & 1.79 \\
& FVMD-ISRe$^{\ast}$ & \textbf{1.57} & 2.81 & 1.10 & 1.72 \\
\midrule
\multirow{3}{*}{3} 
& S2P (MGM) & 4.57 & 5.04 & 1.55 & 2.06 \\
& SkySplat w/o C.A. & 4.24 & 2.68 & 0.90 & 1.53 \\
& SkySplat & 3.86 & \textbf{2.46} & \textbf{0.89} & \textbf{1.51} \\
\bottomrule
\end{tabular}
\caption{\textbf{Effect of the number of views on selected AOIs.} Results marked with $^{\ast}$ are from previous work \cite{zhang2024fvmd}. S2P (MGM) refers to the classic stereo pipeline \cite{de2014automatic}. The evaluation metric is MAE in meters.}
\label{tab:viewnum}
\end{table*}

\begin{table*}[!t]
\centering
\setlength{\tabcolsep}{2.5mm} 
\begin{tabular}{l|ccccc}
\toprule
\textbf{Method} & MFE\,(px)$\downarrow$ & MAE\,(m)$\downarrow$ & RMSE\,(m)$\downarrow$ & ${PAG}_{2.5}$\,(\%)$\uparrow$ & ${PAG}_{7.5}$\,(\%)$\uparrow$ \\
\midrule
HiSplat\,(256×256)     & 0.27 & 13.18 & 16.59 & 15.06 & 37.08 \\
HiSplat\,(512×512)     & 0.43 & 13.96 & 17.68 & 14.62 & 35.71 \\
HiSplat\,(1024×1024)   & 1.23 & 14.04 & 17.77 & 14.49 & 35.48 \\
HiSplat\,(2048×2048)   & 2.76 & 14.36 & 18.11 & 13.84 & 34.40 \\
\midrule
SkySplat\,(RGB sup.)       & ---  & \textbf{6.07} & \textbf{7.33} & \textbf{29.46} & \textbf{68.57} \\
\bottomrule
\end{tabular}
\caption{\textbf{Effect of image size on the DFC19 dataset.} We compare SkySplat (trained with RGB-only supervision) with HiSplat, where all inputs are cropped to 256×256. For HiSplat, the resolutions in parentheses refer to the image sizes used when fitting the pinhole camera models before cropping.}
\label{tab:imagesize}
\end{table*}

\subsection{Ablation Study}
To evaluate the effectiveness of each component, we conduct ablation studies on the DFC19 dataset. Results show that all proposed modules contribute to improved performance.
\subsubsection{Ablation of Cross-Self Consistency Module.} The CSCM is designed to suppress the influence of transient objects by stopping gradient propagation. It improves performance in both settings: without relative height supervision (Row 1 vs. Row 2 in Table \ref{tab:ablation_df19}) and with supervision (Row 3 vs. Row 4). In all cases, adding CSCM yields consistent improvements across metrics.
\subsubsection{Ablation of Relative Height Supervision.} We use relative height from DAMV2 as auxiliary supervision to promote geometry learning. To assess its necessity, we compare results without and with Relative Height Supervision (R.H.S.) (Row 2 vs. Row 4 in Table \ref{tab:ablation_df19}). The results confirm that relative height supervision is essential for better geometric reconstruction. This supervision thus acts as a crucial complementary signal to photometric cues.
\subsubsection{Ablation of Consistency Aggregation.} We disable the C.A. strategy to evaluate its effect on scene refinement (Row 4 vs. Row 5 in Table \ref{tab:ablation_df19}). Removing C.A. leads to significant drops in accuracy, showing that multi-view consistency aggregation is critical for reconstruction. This strategy effectively integrates information from multiple views, reducing noise and enhancing the reliability of refined points.
\subsubsection{Effect of the Number of Views.} While previous methods typically require five input views, our three-view reconstruction approach achieves better results in most cases. As shown in Table \ref{tab:viewnum}, SkySplat outperforms these five-view baselines, indicating that high-quality scene reconstruction can be effectively accomplished with fewer views.
\subsubsection{Effect of Image Size.} To investigate the impact of image size on generalizable 3DGS, we fit pinhole camera models under varying image sizes and report the mean fitting error (MFE) in pixels on the test set (Table \ref{tab:imagesize}). The results highlight two observations: (1) HiSplat shows large errors as fitting size increases, due to poor RPC approximating, which limits its use in large-scale remote sensing. (2) SkySplat, even without auxiliary modules (i.e., supervised using only RGB images), showing greater precision in reconstruction.

\section{Conclusion}
We propose SkySplat, a novel self-supervised framework for generalizable 3D reconstruction from multi-temporal sparse satellite images. Extensive experiments show that SkySplat is up to 86× faster than SOTA per-scene optimization methods, while maintaining strong cross-dataset generalization. By explicitly avoiding per-scene optimization and ground-truth height supervision, SkySplat makes significant progress toward efficient satellite-based 3D reconstruction.
\subsubsection{Limitations.} Our method relies on MVS for height estimation, inheriting its limitations in low-texture or reflective areas, which reduces reconstruction quality. Moreover, DFC19 is currently the only large open-source dataset with multi-temporal satellite images, its limited diversity hinders generalization of SkySplat. This highlights the need for larger, more diverse datasets and improved generalization strategies.

\section{Appendix-A. More EXPERIMENTS}

\subsection{A.1. Hyperparameter Analysis}
Table \ref{tab:hyperparams} analyzes the sensitivity of two hyperparameters: the iteration to activate the CSCM module and the similarity threshold for transient masking. Activating CSCM too early (e.g., 20k) causes higher error due to inaccurate height estimates, while activating too late (e.g., 100k) also degrades performance due to prolonged interference from transient objects. The best result (MAE = 2.07 m) occurs with activation at 35k iterations and a threshold of 0.2. Notably, a too loose threshold (e.g., 0.8) also harms accuracy, emphasizing the need for both timely activation and a proper threshold choice.

\begin{table}[H]
\centering
\setlength{\tabcolsep}{3.5mm} 
\begin{tabular}{c|c|c}
\toprule
iters & thre & MAE (m) ↓ \\
\midrule
0k    & 0.0 & 2.25 \\
20k   & 0.2 & 2.12 \\
35k   & 0.8 & 2.18 \\
100k  & 0.2 & 2.12 \\
35k   & 0.2 & \textbf{2.07} \\
\bottomrule
\end{tabular}
\caption{\textbf{Hyperparameter sensitivity analysis on the DFC19 dataset.} The parameter \textit{iters} indicates when the CSCM module activates (in iterations), and \textit{thre} is the similarity threshold for detecting transient objects.}
\label{tab:hyperparams}
\end{table}

\subsection{A.2. Novel View Synthesis Results}
\begin{figure*}[h!]
    \centering
    \includegraphics[width=\linewidth]{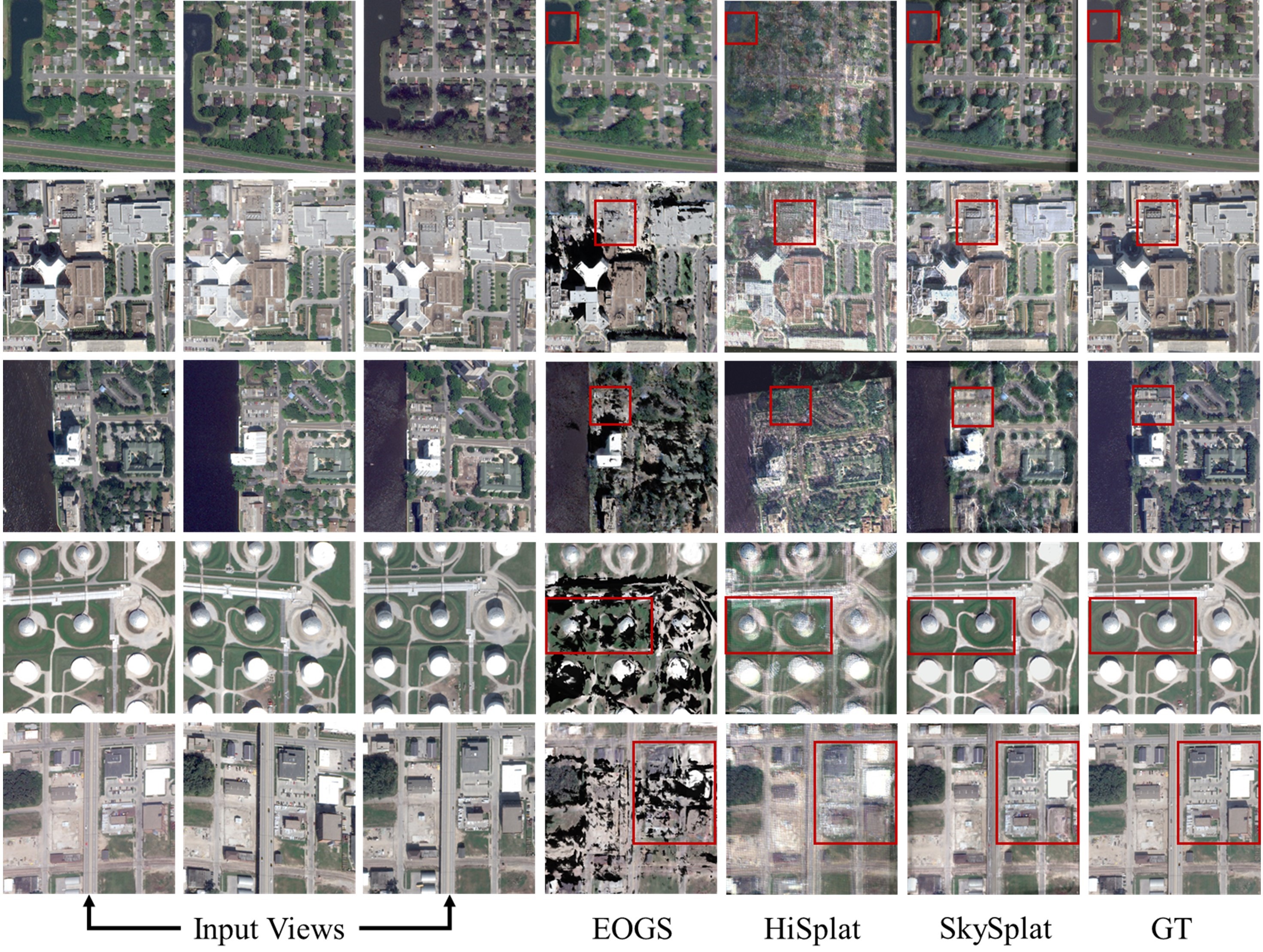}
    \caption{\textbf{Visualization of novel view synthesis on the DFC19 dataset.} From top to bottom: JAX004, JAX068, JAX260, OMA212, OMA315.}
    \label{fig:PSNR}
\end{figure*}

To evaluate the novel view synthesis quality of our model, we approximate the RPC model as the pinhole camera model for rendering novel views, as the error is negligible when the image size is relatively small. We then compare it with the SOTA generalizable 3DGS method HiSplat \cite{tang2024hisplat} and the SOTA per-scene optimization method EOGS \cite{aira2025gaussian} in Table \ref{tab:psnr_lpips_results}, using the Peak Signal-to-Noise Ratio (PSNR) \cite{wang2004image} and the Perceptual Distance (LPIPS) \cite{zhang2018unreasonable} as the evaluation metrics . The results show that our approach consistently achieves superior performance in most cases. We further provide qualitative visualizations in Figure \ref{fig:PSNR}.

\begin{table*}[!t]
\centering
\setlength{\tabcolsep}{2.0mm} 
\begin{tabular}{l|ccccc|ccccc}
\toprule
\multirow{2}{*}{\textbf{Method}} & \multicolumn{5}{c|}{PSNR $\uparrow$} & \multicolumn{5}{c}{LPIPS $\downarrow$} \\
 & JAX004 & JAX068 & JAX260 & OMA212 & OMA315 & JAX004 & JAX068 & JAX260 & OMA212 & OMA315 \\
\midrule
EOGS     & \textbf{22.76} & \underline{14.02} & \underline{13.47} &  8.95 & 10.67 & \textbf{0.399} & \underline{0.539} & \underline{0.612} & 0.649 & \underline{0.631} \\
HiSplat  & 15.25 & 12.60 & 12.64 & \underline{15.76} & \underline{17.81} & \underline{0.698} & 0.678 & 0.663 & \underline{0.625} & 0.641 \\
SkySplat  & \underline{18.87} & \textbf{16.01} & \textbf{16.80} & \textbf{20.74} & \textbf{20.27} & \textbf{0.399} & \textbf{0.463} & \textbf{0.533} & \textbf{0.333} & \textbf{0.382} \\
\bottomrule
\end{tabular}
\caption{\textbf{Quantitative comparison of novel view synthesis performance on five AOIs.} Metrics include PSNR (higher is better) and LPIPS (lower is better). (\textbf{Bold} indicates best, \underline{underline} indicates second best.)}
\label{tab:psnr_lpips_results}
\end{table*}

\subsection{A.3. Effect of Water Mask}
As shown in Table \ref{tab:w_wo_water_mask}, methods such as S-NeRF \cite{derksen2021shadow}, Sat-NeRF \cite{mari2022sat}, and EOGS \cite{aira2025gaussian} rely on the assumption of strictly Lambertian surface reflectance. This assumption breaks down in water regions \cite{zhang2024satensorf}, leading to degraded performance. Excluding water areas from MAE computation (i.e., applying a water mask) alleviates this issue and improves their results. In contrast, SkySplat explicitly models water surfaces, achieving consistently superior performance with or without the mask.

\begin{table*}[!t]
\centering
\setlength{\tabcolsep}{1.7mm} 
\begin{tabular}{l|ccc|cc|ccc|c}
\toprule
\textbf{Method} & JAX004 & JAX068 & JAX260 & OMA212 & OMA315 & IARPA001 & IARPA002 & IARPA003 & Time \\
\midrule
\multicolumn{10}{c}{\textbf{\textit{With Water Mask}}} \\
NeRF       & 3.35 & 6.33 & 3.46 & 1.16 & 3.01 & 4.11  & 6.05  & 5.83  & 5.76 h \\
S-NeRF     & 3.29 & 7.47 & 4.91 & 3.24 & 2.98 & 4.97  & 9.71  & 6.87  & 6.62 h \\
Sat-NeRF   & 3.18 & 6.53 & 5.09 & 3.16 & 2.99 & 4.63  & 6.65  & \underline{4.99}  & 7.39 h \\
EOGS       & \underline{2.57} & 6.67 & 5.00 & 9.08 & 6.38 & 5.90  & 13.79 & 14.34 & 4.60 min \\
SkySplat w/o C.A.   & \textbf{1.66} & \underline{4.24} & \underline{3.14} & \underline{0.90} & \underline{1.53} & \underline{3.14}  & \underline{3.89}  & \textbf{3.25}  & \textbf{3.13 s} \\
SkySplat   & \textbf{1.66} & \textbf{3.86} & \textbf{3.00} & \textbf{0.89} & \textbf{1.51} & \textbf{3.10}  & \textbf{3.75}  & \textbf{3.25}  & 3.19 s \\
\midrule
\multicolumn{10}{c}{\textbf{\textit{Without Water Mask}}} \\
NeRF       & 3.30 & 6.33 & 3.09 & 1.16 & 3.01 & 4.11  & 6.05  & 6.02  & 5.76 h \\
S-NeRF     & 3.28 & 7.47 & 4.88 & 3.24 & 2.98 & 4.97  & 9.71  & 6.55  & 6.62 h \\
Sat-NeRF   & \underline{3.27} & 6.53 & 5.28 & 3.16 & 2.99 & 4.63  & 6.65  & \underline{4.92}  & 7.39 h \\
EOGS       & 3.31 & 6.67 & 6.41 & 9.08 & 6.38 & 5.90  & 13.79 & 14.83 & 4.60 min \\
SkySplat w/o C.A.   & \textbf{1.56} & \underline{4.24} & \underline{2.68} & \underline{0.90} & \underline{1.53} & \underline{3.14}  & \underline{3.89}  & \textbf{3.41}  & \textbf{3.13 s} \\
SkySplat   & \textbf{1.56} & \textbf{3.86} & \textbf{2.46} & \textbf{0.89} & \textbf{1.51} & \textbf{3.10}  & \textbf{3.75}  & \textbf{3.41}  & \underline{3.19 s} \\
\bottomrule
\end{tabular}
\caption{\textbf{Extended comparison with per-scene optimization methods across three cities.}  Reported metrics include MAE (meters) and reconstruction time, both with and without applying the water mask. (\textbf{Bold} indicates best, \underline{underline} indicates second best.)
}
\label{tab:w_wo_water_mask}
\end{table*}

\section{Appendix-B. More Visual Results}
\subsection{B.1. More Results for Height Estimation}
\begin{figure*}[h!]
    \centering
    \includegraphics[width=0.85\linewidth]{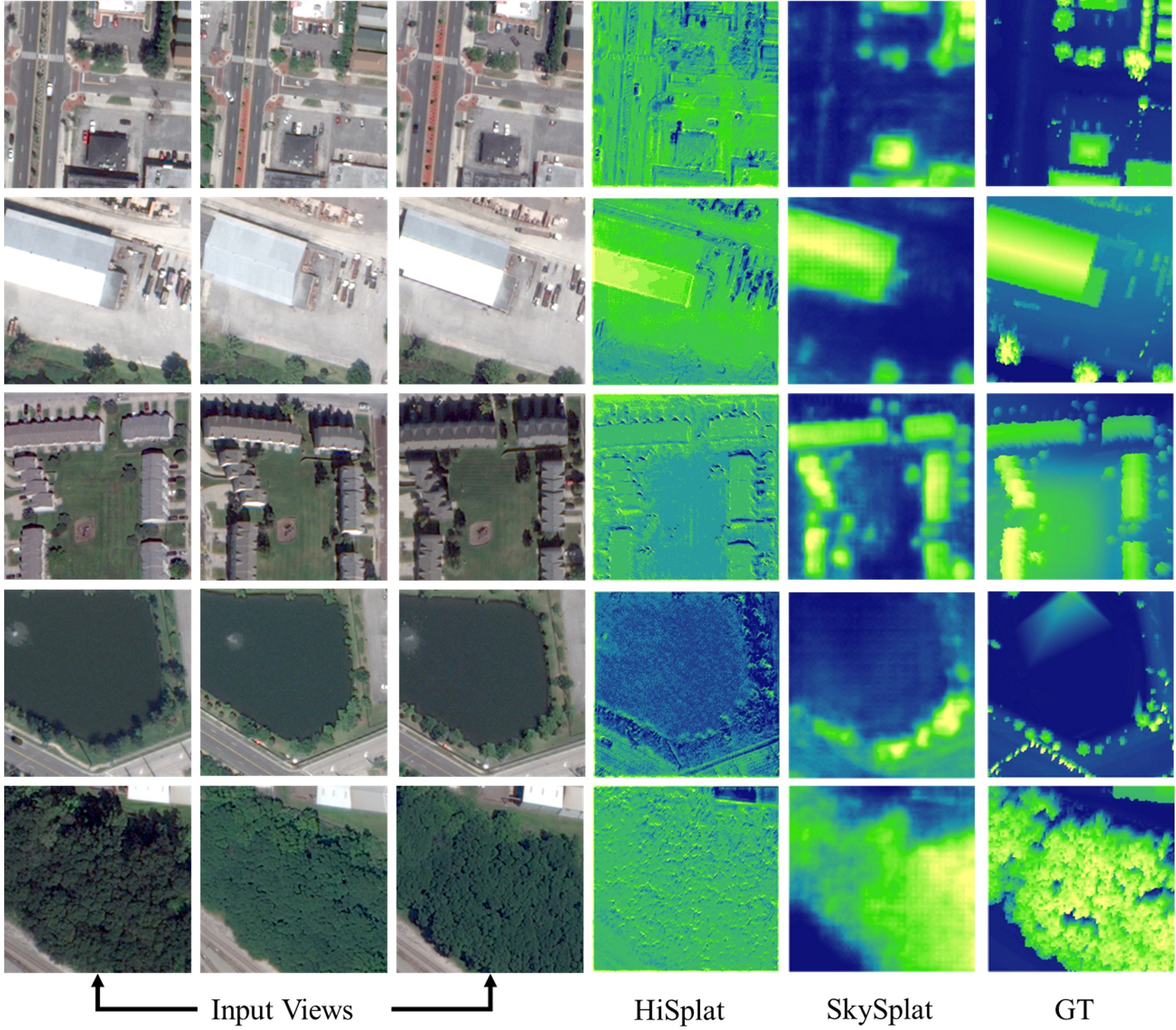}
    \caption{\textbf{Visualization of height predictions on the DFC19 dataset.}}
    \label{fig:Height_Estimation}
\end{figure*}

Figure \ref{fig:Height_Estimation} presents additional visual comparisons of height estimation between SkySplat and HiSplat \cite{tang2024hisplat}. The selected regions cover diverse scene types, including urban, industrial, residential, water, and forested areas. Across all scenarios, SkySplat consistently delivers higher reconstruction quality, highlighting its strong generalization and robustness. For a clearer visual comparison of height estimation, all generalizable 3DGS baselines in this paper are visualized with depth maps (i.e., the distance from the camera to the ground surface).

\subsection{B.2. More Results for 3D Gaussians}
\begin{figure*}[h!]
\centering
\includegraphics[width=0.7\linewidth]{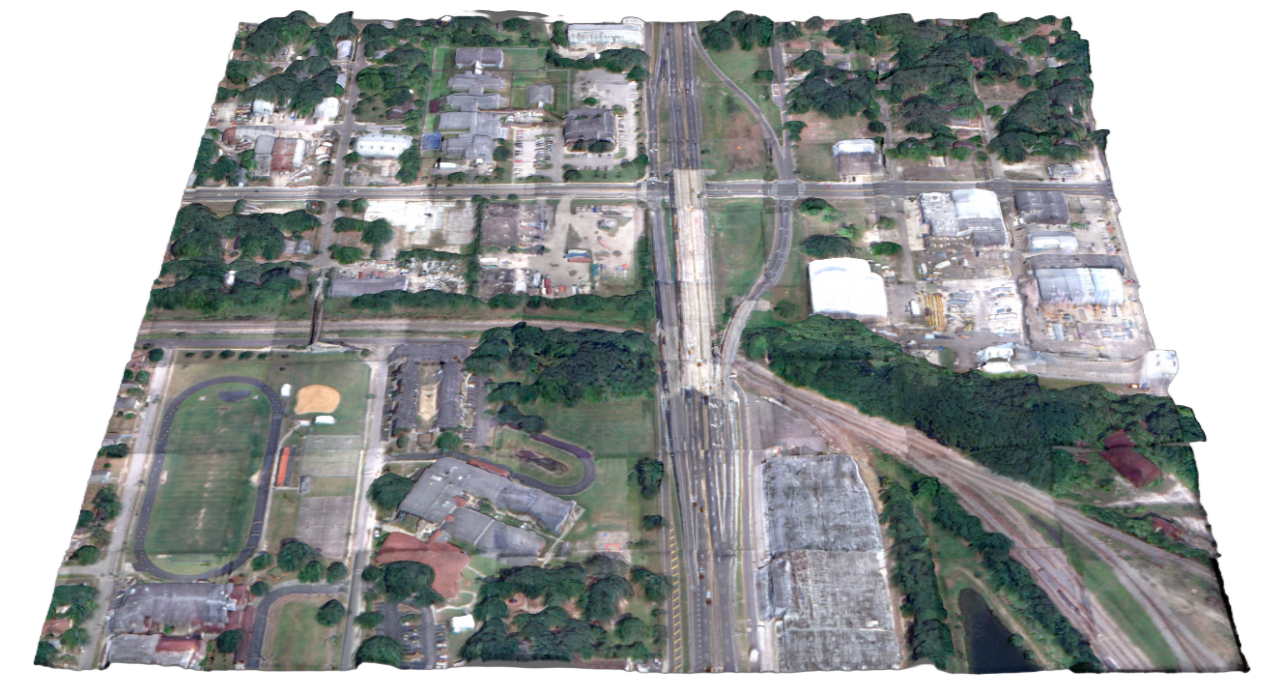}
\caption{\textbf{Visualization of 3D Gaussians on JAX 079.}
}
\label{fig:JAX079}
\end{figure*}

\begin{figure*}[h!]
\centering
\includegraphics[width=0.7\linewidth]{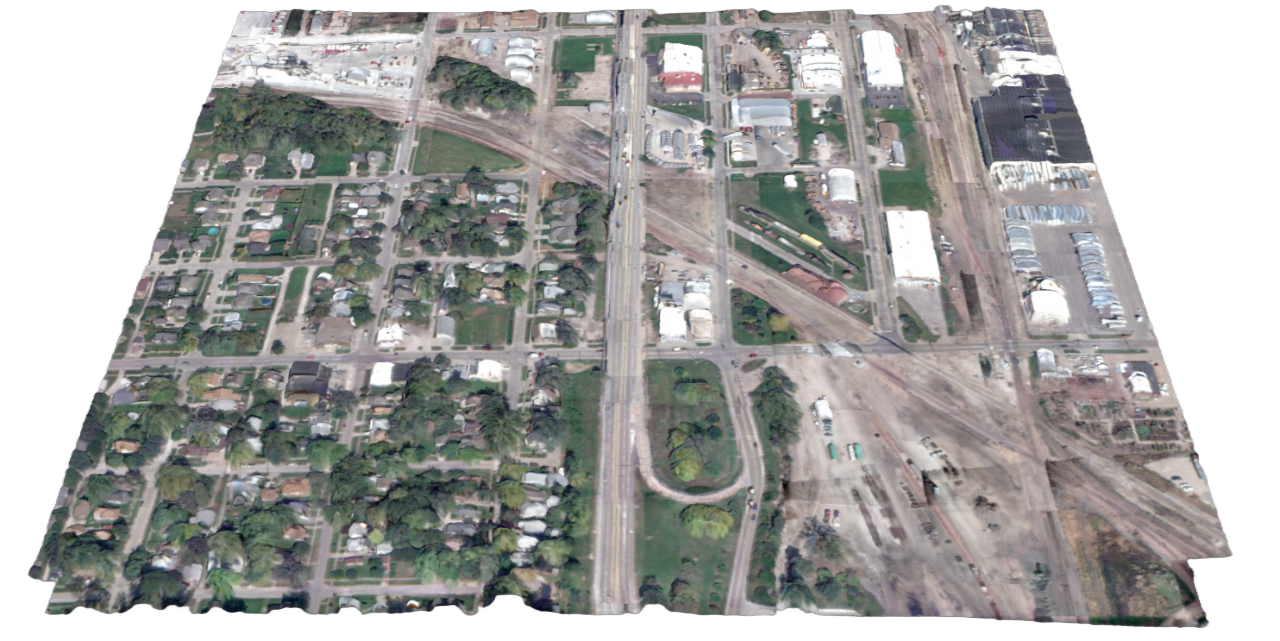}
\caption{\textbf{Visualization of 3D Gaussians on OMA 353.}
}
\label{fig:OMA353}
\end{figure*}

Figures \ref{fig:JAX079}-–\ref{fig:OMA353} showcase more 3D Gaussian results generated by SkySplat. These visualizations demonstrate that our approach consistently achieves accurate and high-fidelity scene reconstructions across diverse scenarios.

\begin{figure*}[h!]
\centering
\includegraphics[width=0.7\linewidth]{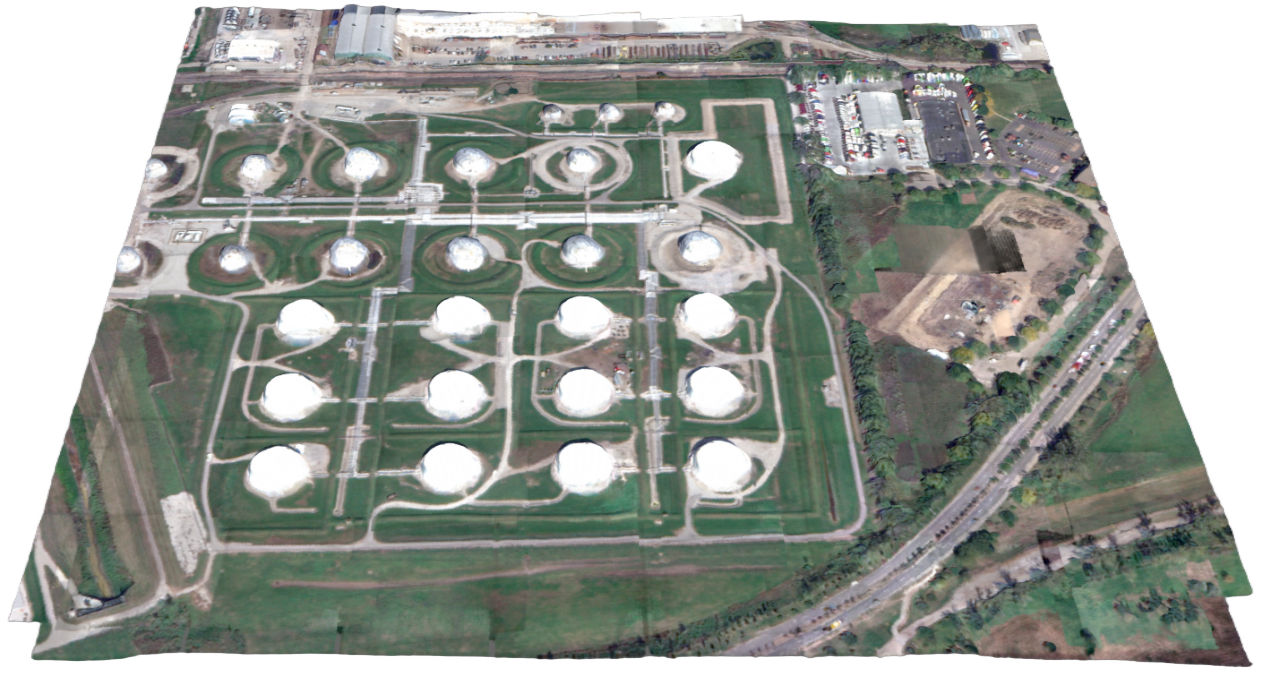}
\caption{\textbf{Visualization of 3D Gaussians on OMA 212.}
}
\label{fig:OMA 212}
\end{figure*}

\section{Appendix-C. More Implementation Details}
We provide additional details on the comparison experiments for the generalizable 3DGS methods. As mentioned in the main text, the RPC model is approximated by the pinhole camera model \cite{zhang2019leveraging}, and higher numerical precision is employed to prevent numerical instability that may arise from large depth values. To ensure a fair comparison, the depth sampling range in the compared methods is aligned with ours. Specifically, the height values from our sampling range are projected to image space using the fitted pinhole camera model, resulting in the corresponding depth sampling range. Furthermore, all hyperparameters in the comparative experiments are kept at their original settings. 

It is important to note that all depth estimation results from the comparison experiments follow the DepthSplat \cite{xu2025depthsplat}; that is, they are obtained from the depth estimation of the network rather than through 3DGS rendering.

\FloatBarrier
\clearpage
\bibliography{aaai2026}
\end{document}